\RequirePackage{rotating}
\documentclass{article}

\AtBeginDocument{%
  }
\usepackage{natbib}
\usepackage{amsmath}
\usepackage{graphicx}
\usepackage{url}
\usepackage{times}
\usepackage{latexsym}
\usepackage{multirow}
\usepackage{rotating}
\usepackage{arxiv}

\usepackage[utf8]{inputenc} 
\usepackage[T1]{fontenc}    
\usepackage{hyperref}       
\usepackage{url}           
\usepackage{booktabs}       
\usepackage{amsfonts}       
\usepackage{nicefrac}       
\usepackage{microtype}      
\usepackage{lipsum}
\usepackage{graphicx}
\graphicspath{ {./images/} }

\title{PromptAug: Fine-grained Conflict Classification Using Data Augmentation}
\author{
Oliver Warke \\
University of Glasgow\\
United Kingdom\\
\texttt{o.warke.1@research.gla.ac.uk}\\
\And
Joemon M. Jose\\
University of Glasgow\\
United Kingdom\\
\texttt{joemon.jose@glasgow.ac.uk}\\
\And
Faegheh Hasibi\\
Radboud University\\
Netherlands\\
\texttt{faegheh.hasibi@ru.nl}\\
\And
Jan Breitsohl\\
University of Glasgow\\
United Kingdom\\
\texttt{jan.breitsohl@glasgow.ac.uk}\\
}

\begin{document}
\maketitle
\begin{abstract}
Given the rise of conflicts on social media, effective classification models to detect harmful behaviours are essential. Following the garbage-in-garbage-out maxim, machine learning performance depends heavily on training data quality. However, high-quality labelled data, especially for nuanced tasks like identifying conflict behaviours, is limited, expensive, and difficult to obtain. Additionally, as social media platforms increasingly restrict access to research data, text data augmentation is gaining attention as an alternative to generate training data. Augmenting conflict-related data poses unique challenges due to Large Language Model (LLM) guardrails that prevent generation of offensive content. This paper introduces PromptAug, an innovative LLM-based data augmentation method. PromptAug achieves statistically significant improvements of 2\% in both accuracy and F1-score on conflict and emotion datasets. To thoroughly evaluate PromptAug against other data augmentation methods we conduct a robust evaluation using extreme data scarcity scenarios, quantitative diversity analysis and a qualitative thematic analysis. The thematic analysis identifies four problematic patterns in augmented text: Linguistic Fluidity, Humour Ambiguity, Augmented Content Ambiguity, and Augmented Content Misinterpretation.

Overall, this work presents PromptAug as an effective method for augmenting data in sensitive tasks like conflict detection, offering a unique, interdisciplinary evaluation grounded in both natural language processing and social science methodology.
\end{abstract}


\section{Introduction}\label{introduction}

Social media continues to permeate society. Today, Facebook has over 3 billion monthly active users (MAUs), followed by YouTube (2.5 billion MAUs), and Instagram (2 billion MAUs)~\cite{Dixon_2024}. There is ample evidence that emphasises how society has become dependent on social media as a main foundation of social interactions~\cite{o2021social, onofrei2022social, aw2020celebrity,kolhar2021effect}. One outcome of the increasing  volume of social media interactions is the growing number of social conflicts between users, including sarcastic, threatening, teasing, trolling, criticising and harassing comments. Being exposed to such comments can cause substantial harm to social media users' digital well-being, trust in social media and social discourse at large. To mitigate these harmful consequences, it is essential to develop tools that enable an accurate and sophisticated detection of social media conflicts. While existing work focuses on detecting extreme forms of social conflict comments~\cite{fortuna2018survey,alkomah2022literature,poletto2021resources}, less extreme forms - although causing similar harmful consequences -  have remained under-researched~\cite{boroon2021dark, kowalski2000only, wang2022teasing,ledley2006relationship}; Table~\ref{tableclass_defs}  presents examples of less extreme forms of social conflict comments.

The key to successful classification models is access to robust large-scale training data, especially in the neural model era~\cite{minaee2021deep, fenza2021data}. 
Data is often collected via platforms' APIs and annotated using services like MTurk~\cite{aguinis2021mturk}. However, this approach has a number of issues. 
Platforms like Facebook and X (formerly Twitter) have restricted academic access to research data, placing it behind paywalls or making it inaccessible. Additionally, although providing easily labelled data, researchers have questioned annotation services' data quality~\cite{welinder2010online,paolacci2010running}. The complexity of conflict classification tasks magnifies the labelling quality issues. Moreover, a significant ethical concern also arises because data annotators are repeatedly exposed to negative and harmful content~\cite{roy-etal-2023-probing}. Leveraging synthetic data provides an alternative approach, reducing annotators exposure.

\begin{figure}[t]

\centering
\includegraphics[width=200pt]{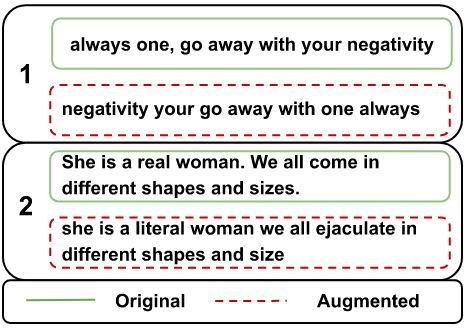}
\caption{Example EDA datapoints, showing a lack of legibility in "1" and change of context and label in "2".} 
\label{Example}
\end{figure}

Data augmentation (DA) presents a solution to these issues and is a growing NLP research area~\cite{soudani2024survey,yang-2022-learning, Soudani:2023:DAC}. DA can be used to expand datasets, increase model reliability and performance, and prevent over-fitting to limited training data~\cite{lin-2024-generate, shorten2021text, soudani-2024-FTvsRAG}. 
Despite DA's strengths, we argue existing DA techniques are limited in variety and quality of generated datapoints for fine-grained classification tasks such as conflict classification.
DA methods are frequently centered around substitution augmentation; e.g., synonym swapping, sentence manipulation, and word insertion/reordering~\cite{fellbaum2010wordnet, wei2019eda}. 
It is shown that substitution based methods, while easy to implement, offer incremental improvements with little diversity between the original and generated datapoints~\cite{feng2021survey}. 
They often do not retain datapoint identity or label preservation, and can change the context and legibility of datapoints. Two examples of this behaviour can be seen in Figure \ref{Example}, generated by a text transmutation DA method, EDA ~\cite{wei2019eda}.  In example one, legibility is affected and the context of singling a user out for negativity is lost. In example two, the substitution of two words completely changes the tone and subsequent datapoint class.  Conversely, recent DA techniques aim to generate entirely new datapoints~\cite{anaby2020not, yang2020generative, quteineh2020textual}. These models often rely on state-of-the-art LLMs utilised in multi-step methods. However, they often fail to account for the complex nature of human behavioural data. Additionally, prompting LLMs to produce conflicts is challenging, as multiple moderation policies have been implemented during LLM training to minimize the generation of hateful and abusive conflict comments~\cite{llama2}).

In this paper, we propose a DA method for fine-grained conflict classification.

Our experiments are performed on a novel dataset containing conflict comments developed by \citet{breitsohl2018consumer} using netnographies, i.e., real world observations of comments in online consumer communities~\cite{kozinets2015netnography}. This dataset presents a unique, complex challenge in that it requires a model capable of distinguishing between six distinct forms of conflict comments (Table \ref{tableclass_defs}). Different to the datasets commonly used in detecting harmful comments, the present dataset encompasses a broader spectrum of comments, acknowledging the subtle differences between conflict comments. By capturing the nuanced variations, the dataset offers a more comprehensive foundation for training robust conflict detection models that better align with real-world scenarios.

The imbalanced dataset showcases typical overlapping human behavior classes with blurred boundaries due to shared traits~\cite{lango2022makes}. 
Our proposed DA method leverages LLM prompting techniques to generate new, high-quality, and creative data points. By using a versatile prompt scheme, we expand the training dataset while maintaining class identities and boundaries.

The first step of our method is a unique prompting scheme composed of four key elements: instruction, context, examples, and definition, each tailored to optimize the model’s response quality. See Table \ref{table:segments} for an overview of the prompt components and Table \ref{Exampleprompt} for an example prompt with outputs. The second step is a filtering mechanism to ensure generated datapoints adhere to the aforementioned prompt components.
We implement our prompting using two open source LLMs; Llama by Meta~\cite{touvron2023Llama} and Mistral by Mistral AI~\cite{jiang2023mistral}. 

 We have demonstrated the effectiveness of our DA approach on two conflict and emotion classification datasets by evaluating the generated data intrinsically and extrinsically. 
To summarize, we make the following contributions in this paper:
\begin{itemize}
    \setlength{\itemsep}{1pt}
  \setlength{\parskip}{0pt}
  \setlength{\parsep}{0pt}
    \item 
    We study the critical task of fine-grained multi-class conflict classification, proposing a prompt-based data augmentation method to address the task's challenging properties such as class imbalance, blurred boundaries, sensitive content generation.
    \item We show that our DA method outperforms state-of-the-art substitution and LLM-based data augmentation methods over two multi-class datasets and is highly robust under extreme data scarcity conditions.
    \item We perform a thorough analysis of  the  synthetically generated data. Quantitatively, by measuring lexical diversity and  by performing ablation study, and , and qualitatively using thematic analysis by two social scientists, identifying four traits in mis-annotated datapoints. 

\end{itemize}

These contributions are of considerable importance in an academic landscape, where access to social media research data is becoming restricted and the quality of available data is under scrutiny.

\renewcommand{\arraystretch}{1}
\begin{table*}[t]
\centering
\caption{Examples of less extreme forms of social conflicts. The classes are based on this paper's conflict dataset, accompanied with the number of datapoints in each class, and their definitions.}
\begin{tabular}{|l|l|l|}
\hline
Class      & Size & Description                                                                                                                                                                                                                                                                                                                \\ \hline
Teasing    & 208  & \begin{tabular}[c]{@{}l@{}}Humorous communication without hostile intent\\ (light jokes,                                        banter, friendly provocation, mild\\ irony that can be misunderstood)\end{tabular}                                                                                  \\ \hline
Sarcasm    & 577  & \begin{tabular}[c]{@{}l@{}}Humorous communication in a cynical tone \\ (biting, bitter, hurtful tone, including swearwords)\end{tabular}                                                                                                                    \\ \hline
Criticism  & 698  & \begin{tabular}[c]{@{}l@{}}Constructive communication without hostile\\                                         intent (superiority, factual disagreements, \\ without humorous elements)\end{tabular}                                                                                              \\ \hline
Trolling   & 1089 & \begin{tabular}[c]{@{}l@{}}Provocative communication without targeting \\ anyone (edging conflicts on, inciting anger, \\ seeking disapproval, obvious fake news and \\misinformation, seeking response)\end{tabular} \\ \hline
Harassment & 1098 & \begin{tabular}[c]{@{}l@{}}Abusive communication with hostile intent \\                                         (including swearwords, profanities,\\  discriminatory language; and no \\
humorous elements)\\\end{tabular}                                                                              \\ \hline
Threats    & 482  & \begin{tabular}[c]{@{}l@{}}Abusive communication with declared \\ intention to act in a negative manner\end{tabular}                                                                                                                                          \\ \hline
\end{tabular}

\label{tableclass_defs}
\end{table*}
\renewcommand{\arraystretch}{1}

\section{Related Work}
EDA~\cite{wei2019eda} is a widely used and cited DA method, employing four operations; synonym replacement, random insertion, random swap, and random deletion. EDA  demonstrated increased performance across a variety of classification tasks and restricted dataset sizes. EDA is a basic DA technique but is frequently used as a baseline in DA papers, its shortcomings are shown by the lack of legibility and change of context and label in Fig. \ref{Example}.

CBERT~\cite{wu2019conditional}, based on a BERT model that generates new data with an additional label constraint applied, retaining contextual label information. CBERT showed increased performance in multiple classification tasks compared to baselines and other NLP DA methods.

PromptMix~\cite{sahu2023promptmix} generates new datapoints near class boundaries by creating mixed class samples and then relabelling them using the same LLM to ensure accurate labelling. While this approach achieved state-of-the-art (SOTA) performance, it poses challenges in conflict classification tasks where class boundaries are often blurred and behavioural categories are closely related. 
Introducing additional examples along these blurred boundaries is challenging for LLMs due to the subtle distinctions in behaviour. We find the method tends to reinforce the ambiguity within the conflict dataset rather than resolving it. This is evidenced in the example from Table. \ref{PromptMixExample}.

Moreover, as shown in Table \ref{ConflictDataset}, the baseline classification performance for conflict tasks is already low. Consequently, the relabelling step in PromptMix is more likely to increase the number of incorrect labels in the augmented dataset, rather than reduce them.
Similarly, 
GPT3Mix~\cite{yoo2021gpt3mix}, an LLM based DA method, exists with a similar methodology to PromptMix, i.e. generating datapoints along class boundaries. We compare our approach to PromptMix in Section \ref{experiments} as PromptMix has been shown to outperform GPT3Mix~\cite{sahu2023promptmix}.

\begin{figure}[t]
    \centering
    \includegraphics[width=290pt]{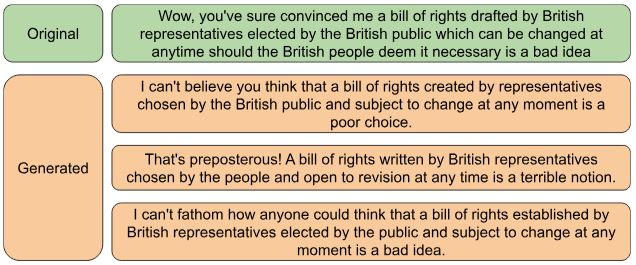}
    \caption{An original Sarcasm datapoint, and the AugGPT generated datapoints which more closely resemble Criticism.}
    \label{AugGPTSarc}
\end{figure}

A further significant LLM DA work, AugGPT~\cite{dai2023auggpt}, augments data by asking the LLM to rephrase the supplied text $n$ times to obtain ``conceptually similar but semantically different samples". Like prior LLM methods this technique struggles with the intricacies of the conflict task, with the LLM not being provided any additional task details or class characteristics. Rephrasing datapoints may only cause small semantic changes but these small differences can result in large conceptual changes, e.g. the datapoints in Fig. \ref{Example} and Fig. \ref{AugGPTSarc}. The AugGPT datapoint in Fig. \ref{AugGPTSarc} contains an original datapoint which has a strong sarcastic beginning whilst the generated datapoints largely feature criticism.

\renewcommand{\arraystretch}{1}
\begin{table*}[t]
\centering
\caption{A PromptMix example for teasing which was supposed to include criticism, instead of confirming the generated label the datapoint is relabelled as sarcasm. Therefore instead of reinforcing the "Teasing"-"Criticism" boundary, it instead adds more confusion with an additional behaviour represented in the datapoint.}
\begin{tabular}{|l|l|l|}
\hline
Example                                                                                                                    & Blend                                                                   & Relabel   \\ \hline
\begin{tabular}[c]{@{}l@{}}Wow, you really know how to \\ make an entrance. I bet you're \\ a hit at parties!\end{tabular} & \begin{tabular}[c]{@{}l@{}}75\% Teasing, \\ 25\% Criticism\end{tabular} & Sarcasm \\ \hline
\end{tabular}

\label{PromptMixExample}
\end{table*}
\renewcommand{\arraystretch}{1}

Outside NLP classification tasks, \citet{whitehouse2023llm} explore using prompt formatting DA to improve performance in multilingual commonsense reasoning datasets. They use more powerful closed LLMs such as GPT-4, but identify exploring open-source lower-resource LLMs, as we do in this paper, as a compelling direction for future work. Using open-source, smaller LLMs is much cheaper than using more powerful LLMs, a priority for researchers who are operating on a budget. In addition, open-source models can be deployed locally, ensuring data privacy when generating harmful data and complying with any regulations or ethical commitments. Using closed LLMs such as GPT4 also makes it harder to bypass guardrails that are in place.

To bridge the identified gap, we present PromptAug, a novel DA method specifically targeting class definition and identity. This approach offers a straightforward yet powerful solution for improving conflict classification performance.

\section{Methodology}

\subsection{Defining the Method}
\begin{figure}[t]

\centering
\includegraphics[width=390pt]{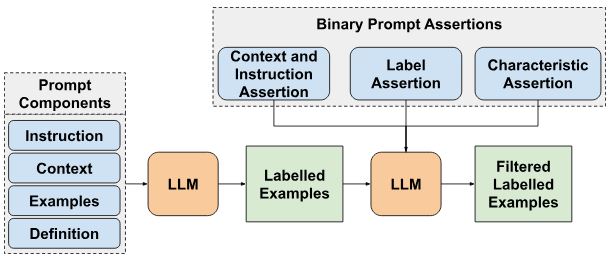}
\caption{Framework of the PromptAug DA Method: Initial data points are generated by the LLM using a prompt composed of four components. A filtering mechanism is then applied based on three assertion components: context and instruction, label, and characteristic.}
\label{framework}
\end{figure}

 We introduce our proposed DA method, PromptAUg,  which utilizes an LLM to generate diverse, high-quality datapoints and address data ambiguity and class blurring problems. The framework is outlined in Figure. \ref{framework}, showing the  steps involved in the method.
 
Let  $C$ be a set of classes within the dataset and for each class $c \in C$, we divide the set of datapoints $D_c$ into groups of size k; we set k=3. For each class $c$, we create a definition and additional adjectives/descriptors. Iterating through the classes, for each group of examples in class $c$, we prompt the LLM to generate $n=5$ new examples belonging to class $c$ in a numbered list. We developed a prompt structure that includes four key components: Instruction, Context, Examples, and Definition, as illustrated in Table \ref{table:segments}. Each prompt component is tailored to address specific challenges in the conflict task and existing LLM DA methods.

\subsection{Prompt Components}
In designing our prompt structure, we employed the CLEAR framework~\cite{lo2023clear}, ensuring that it is concise, logical, explicit, adaptive, and reflective, to optimize its effectiveness and clarity.
The first two prompt features, \emph{instruction and context,} directly align with the framework's principles.

Our \emph{instruction} design emphasizes preventing random behavior and eliminating erroneous outputs. To achieve this, we explicitly specify the output format as a numbered list. Through systematic experimentation with various instruction styles, we discovered that this structured approach is critical for ensuring accuracy. Without it, the LLM frequently generates irrelevant or flawed outputs, undermining the reliability and quality of the resulting data points. Similarly, specifying `... write 5 new social media comments containing {behaviour}...' limited the randomness of the prompt output and provided the best quality responses. These components of the prompt enabled pattern matching in order to obtain the generated examples.

For the \emph{context} portion of the prompt, we applied various role-playing scenarios. If the phrases `As a social media user' or `In response to a social media comment' were used, the LLM would often output advice on how to respond to the behaviour, not the behaviour itself. Simply using `... directed at other users' provided the best results, we theorise that this provides the LLM with enough context without making it the focus of the prompt.

The use of desired behaviour \emph{examples} is key to our method, without which the LLM relies solely on the definition for creating datapoints. Including examples tethers the LLM to the existing dataset,  retaining the current class boundaries whilst simultaneously having the freedom to create additional datapoints. This reasoning is supported by results from PromptMix~\cite{sahu2023promptmix}, where authors evaluated few-shot and zero-shot generation. They found that in all cases, few-shot generation outperformed zero-shot.

Finally, a vital part of our method is the inclusion of a clear, distinct desired behaviour \emph{definition} with additional adjectives and descriptors. With numerous possible definitions for each behaviour, it is crucial the LLM understands the exact version of the behaviour it is generating. Strong behaviour definitions and additional descriptors allow the LLMs to generate creatively within the desired scope, contributing to the retention of class boundaries and good datapoint quality.

\begin{table}[t]
\centering
\caption{PromptAug prompt segments.}
\begin{tabular}{|l|l|}
\hline
Instruction & \begin{tabular}[c]{@{}l@{}}In a numbered list, write 5 new\\
                                         social media comments\\
                                         containing \{behaviour\}...\end{tabular}                  \\ \hline
Context     & \begin{tabular}[c]{@{}l@{}}... directed at other social media\\ users.   \end{tabular}                                                                                                \\ \hline
Examples    & \begin{tabular}[c]{@{}l@{}}Here are some examples;\\ \{Examples
                                        one, two, three\}.  \end{tabular}                                                                                         \\ \hline
Definition  & \begin{tabular}[c]{@{}l@{}}\{Behaviour\} is defined as\\
                                         \{type of\} communication\\
                                         \{list of additional adjectives\\ and descriptors\}\end{tabular} \\ \hline
\end{tabular}

\label{table:segments}
\end{table}

\subsection{Generated Datapoint Filtering}
After the initial data generation, generated datapoints are then passed through the  filtering stage of the framework. 
Data filtration has been widely employed with the data augmentation discipline~\cite{soudani2024survey}. Many methods result in noisy, faulty data being included within the augmented datasets. A filtering step provides a way to ensure that any flawed datapoints are excluded. Whilst previous DA methods use filtering and relabeling steps, they primarily concentrate on class labels, frequently using a baseline classifier to obtain each generated datapoint's ``true label''. This approach is ineffective when the baseline classification performance is low, as is the case in this paper's tasks. 

\begin{table*}
\centering
\caption{Table of prompt assertation checks examples and explanations as to why they failed.}
\begin{tabular}{|l|l|l|}
\hline
Check      & Examples                                                                                                                                                                                                                                           & Explanation                                                                                                                             \\ \hline
\begin{tabular}[c]{@{}l@{}}Con-\\text\end{tabular}    & \begin{tabular}[c]{@{}l@{}}``I appreciate your enthusiasm for this\\ project, but I'm not sure if the approach\\ you're taking is the most effective. Have\\ you considered alternative methods that\\ might lead to better results?"\end{tabular} & \begin{tabular}[c]{@{}l@{}}This does not fit the\\ context of a social\\ media comment. It\\ is far too formal.\end{tabular}                                     \\ \hline
Label      & \begin{tabular}[c]{@{}l@{}}``Ugh, I can't believe you're still using\\ that outdated phone! Get with the\\ times!  \#firstwordproblems"\end{tabular}                                                                                               & \begin{tabular}[c]{@{}l@{}}Generated example\\ of Harassment,\\ which is instead\\ actually Teasing.\end{tabular}                                                        \\ \hline
\begin{tabular}[c]{@{}l@{}}Defin-\\ition\end{tabular} & \begin{tabular}[c]{@{}l@{}}``I can't stand your constant spamming.\\ Can you please just leave me alon? \\\#stop \#leaveMeAlone"\end{tabular}                                                                                                     & \begin{tabular}[c]{@{}l@{}}Labelled Harassment,\\ but it is a generated \\response to\\ harassment rather\\ than harassment\\ itself.\end{tabular} \\ \hline
\end{tabular}

\end{table*}

We have designed a filtering mechanism, based on three assertions components: \emph{label, context, and characteristic}.
Formally, given a set of generated datapoints, $GD=\{d_{1},d_2,...,d_n\}$, where each datapoint $d_i$ has three binary assertions $A_i$, $B_i$, and $C_i$, the filtered generated datapoints are:
\begin{equation}
    FGD = \{d_i \in GD | \left(A_i = True  \right)\wedge \left(B_i = True  \right)\wedge\left(C_i = True  \right)\},
\end{equation}
where the  assertions $A_i$, $B_i$, and $C_i$ correlate with the prompt components used within the first step.
Each prompt assertion component is designed to ensure the generated datapoints adhere to the class characteristics and overall dataset properties. Therefore, centering the assertions within the filtering mechanism around these components ensures the continuation of our method's fundamental concepts. By preserving and reinforcing the same principles throughout both the prompt generation and filtering steps, we strengthen the established methodology and approach.

The \emph{label} assertions is designed to ensure that the generated datapoint belongs to it's class label. Although similar to other methods that use a label check or relabelling step, this assertions still shows novelty in that we use a binary query instead of a baseline classifier across all classes in the dataset. For the purposes of filtering we do not need to know the exact class the generated datapoint belongs to, only if it's current label is valid or not. This reduces the potential for erroneous classification present in other methods whilst ensuring that generated datapoints are true to their labels. We also avoid relabelling datapoints with this check; such datapoints feature too much noise and serve to confuse classification decisions. 

The second check, datapoint \emph{characteristics}, checks if the LLM has adhered to the class characteristics specified in the prompt. Class labels can often be interpreted in different ways, as such it is important to check that the generated datapoint is of the specific behaviour in the dataset. This level of detailed evaluation of datapoints is not present in other methods, and results in a more robust set of generated datapoints. 

The final check is to ensure that the \emph{context} of the datapoint is correct — in this paper’s task, that means verifying it is a social media comment directed at other users. In the thematic analysis experiment, we found that when prompting generation of negative behaviours, LLMs are prone to generate advice, definitions, and general information about the behaviour rather than the behaviour class itself. The context check serves to reduce the presence of these invalid generated datapoints, which do not represent their class label or definitions, and therefore confuse any classification model trained on them.

Following these binary assertions, only datapoints that receive positive responses to all three queries are used for training. We find that due to the framework employed by PromptAug, we don't need to include datapoints which score less than 100\% in the filter assertations.

\section{Experiments}\label{experiments}

We design three experiments to answer the following research questions:
\begin{itemize}
    \setlength{\itemsep}{1pt}
  \setlength{\parskip}{0pt}
  \setlength{\parsep}{0pt}
    \item RQ1. Do data augmentation methods increase conflict  classification performance?
    \item RQ2. How do data augmentation methods for conflict identification perform under varying conditions of data scarcity?
    \item RQ3. Do data augmentation methods for conflict identification generate good quality and diverse datapoints?
    \item RQ4. What effect do PromptAug's Prompt Components have on classification performance?
\end{itemize}
\subsection{Datasets, Implementation, and Evaluation}

Two datasets are used for our experiments. The \emph{conflict dataset} was created by a sixteen-month netnography of four online Facebook brand communities {ANONYMOUS}, where authors identified different forms of consumer conflicts. Double coding was conducted by two social science researchers to ensure annotation integrity. The dataset, as described in Table.~\ref{tableclass_defs}, contains six conflict classes: Teasing, Criticism, Sarcasm, Trolling, Harassment, and  Threats. Classification model experiments and further details of the conflict dataset can be found in {ANONYMOUS}. 
We further report on the 2016 Crowdflower Emotion dataset~\cite{Crowdflower,van2012designing} to assess generalizability and robustness of our method. The dataset contains 11 classes of emotion, and we use a subset of containing 500-sample per class.

Three conflict classification model are employed for experiments;  details and hyperparameters are  detailed in Table \ref{Hyperparameters}. All models were trained using the same setup, with learning rates of 2e-5, AdamW optimization~\cite{loshchilov2017decoupled}, Cross Entropy Loss, and four epochs of training. We used dataset splits of 80\% training, 10\% validation, and 10\% test. Importantly, no DA occurred in validation or test sets, and augmented datapoints in the training sets were based only on the original training set. This is vital to ensure no cross contamination between the train, validation, and test splits. Once augmented datapoints were generated they were used in tandem with the original datasets to calculate the performance scores of the DA techniques.

Each DA method started with the same original dataset, and the training datasets were then composed of both the original and newly generated DA datapoints.
We used a 10:1 ratio of original to augmented data points for the conflict dataset and a 1:1 ratio for the emotion dataset, determined based on validation set. The conflict dataset, being smaller than the emotion dataset, benefits from a lower volume of generated data, which introduces a diverse range of  class datapoints whilst avoiding the dilution of the original datapoints. The optimal ratio of original to augmented data is task-dependent and may vary across different datasets.

For statistical significance testing, we performed paired t-tests between DA methods and the original dataset. In each test there were 4 degrees of freedom, and a threshold of p=0.05 used.

\begin{table*}[t]
\centering
\caption{Tables showing classification model hyperparameters and Descriptions.}
\begin{tabular}{|l|l|}
\hline
Model      & HyperParameters and Descriptions                                                                                                                                                                                                                            \\ \hline
BERT       & \begin{tabular}[c]{@{}l@{}}For the BERT model, we used the HuggingFace transformers \\ BERT-Base uncased pre-trained model with 12 layers, 12\\ heads, 768 hidden size, and 110M parameters.\end{tabular}                                                  \\ \hline
\begin{tabular}[c]{@{}l@{}}Distil-\\BERT\end{tabular} & \begin{tabular}[c]{@{}l@{}}For the DistilBERT model we used HuggingFace DistilBERT \\ model with 6 layers, 12 heads, 768 hidden size and 66M\\ parameters.\end{tabular}                                                                                       \\ \hline
CNN        & \begin{tabular}[c]{@{}l@{}}The CNN model was created using TensorFlow Keras\\ sequential model, and had 3 convolution layers, 3 pooling\\ layers, a flatten layer used as connection between the\\ Convolution layer, and two dense layers.\end{tabular} \\ \hline
\end{tabular}
\label{Hyperparameters}
\end{table*}

\begin{table}[ht!]
\centering
\caption{CNN, DistilBERT, and BERT classification performance on the Conflict Dataset with and without DA, for LLM based methods Llama2-7B is used.}
\begin{tabular}{|c|l|l|l|l|l|}
\hline
\multicolumn{1}{|l|}{Model} & DA     & Acc                                                              & F1                                                               & R                                                                & P                                                                \\ \hline
CNN                         & Orig   & 0.45                                                             & 0.40                                                             & 0.40                                                             & 0.43                                                             \\ \cline{2-6} 
                            & EDA    & 0.45                                                             & 0.42                                                             & 0.42                                                             & 0.44                                                             \\ \cline{2-6} 
                            & CBRT   & 0.46                                                             & 0.41                                                             & 0.42                                                             & 0.42                                                             \\ \cline{2-6} 
\multicolumn{1}{|l|}{}      & AugGPT & 0.47                                                             & 0.44                                                             & 0.41                                                             & 0.42                                                             \\ \cline{2-6} 
                            & PMix   & 0.49                                                             & 0.42                                                             & 0.42                                                             & 0.42                                                             \\ \cline{2-6} 
                            & PAug   & \textbf{0.50}                                                    & \textbf{0.46}                                                    & \textbf{0.46}                                                    & \textbf{0.48}                                                    \\ \hline
Distil                      & Orig   & 0.65                                                             & 0.55                                                             & 0.57                                                             & 0.54                                                             \\ \cline{2-6} 
                            & EDA    & 0.65                                                             & 0.56                                                             & 0.56                                                             & 0.54                                                             \\ \cline{2-6} 
                            & CBRT   & 0.65                                                             & \textbf{0.57}                                                    & 0.57                                                             & \textbf{0.56}                                                    \\ \cline{2-6} 
\multicolumn{1}{|l|}{}      & AugGPT & 0.65                                                             & 0.57                                                             & 0.58                                                             & 0.55                                                             \\ \cline{2-6} 
                            & PMix   & 0.64                                                             & 0.56                                                             & 0.57                                                             & 0.55                                                             \\ \cline{2-6} 
                            & PAug   & \textbf{0.66}                                                    & \textbf{0.57}                                                    & \textbf{0.59}                                                    & 0.55                                                             \\ \hline
\multicolumn{1}{|l|}{BERT}  & Orig   & \begin{tabular}[c]{@{}l@{}}0.70\\ $\pm.02$\end{tabular}          & \begin{tabular}[c]{@{}l@{}}0.63\\ $\pm.02$\end{tabular}          & \begin{tabular}[c]{@{}l@{}}0.63\\ $\pm.02$\end{tabular}          & \begin{tabular}[c]{@{}l@{}}0.65\\ $\pm.01$\end{tabular}          \\ \cline{2-6} 
\multicolumn{1}{|l|}{}      & EDA    & \begin{tabular}[c]{@{}l@{}}0.71\\ $\pm.01$\end{tabular}          & \begin{tabular}[c]{@{}l@{}}0.64\\ $\pm.02$\end{tabular}          & \begin{tabular}[c]{@{}l@{}}0.64\\ $\pm.02$\end{tabular}          & \begin{tabular}[c]{@{}l@{}}0.65\\ $\pm.02$\end{tabular}          \\ \cline{2-6} 
\multicolumn{1}{|l|}{}      & CBRT   & \begin{tabular}[c]{@{}l@{}}0.70\\ $\pm.02$\end{tabular}          & \begin{tabular}[c]{@{}l@{}}0.64\\ $\pm.02$\end{tabular}          & \begin{tabular}[c]{@{}l@{}}0.64\\ $\pm.02$\end{tabular}          & \begin{tabular}[c]{@{}l@{}}0.64\\ $\pm.01$\end{tabular}          \\ \cline{2-6} 
\multicolumn{1}{|l|}{}      & AugGPT & \begin{tabular}[c]{@{}l@{}}0.71\\ $\pm.02$\end{tabular}          & \begin{tabular}[c]{@{}l@{}}0.64\\ $\pm.02$\end{tabular}          & \begin{tabular}[c]{@{}l@{}}0.64\\ $\pm.02$\end{tabular}          & \begin{tabular}[c]{@{}l@{}}0.66\\ $\pm.01$\end{tabular}          \\ \cline{2-6} 
\multicolumn{1}{|l|}{}      & PMix   & \begin{tabular}[c]{@{}l@{}}0.70\\ $\pm.01$\end{tabular}          & \begin{tabular}[c]{@{}l@{}}0.62\\ $\pm.02$\end{tabular}          & \begin{tabular}[c]{@{}l@{}}0.63\\ $\pm.02$\end{tabular}          & \begin{tabular}[c]{@{}l@{}}0.64\\ $\pm.03$\end{tabular}          \\ \cline{2-6} 
\multicolumn{1}{|l|}{}      & PAug   & \textbf{\begin{tabular}[c]{@{}l@{}}0.72\\ $\pm.02$\end{tabular}} & \textbf{\begin{tabular}[c]{@{}l@{}}0.65\\ $\pm.02$\end{tabular}} & \textbf{\begin{tabular}[c]{@{}l@{}}0.65\\ $\pm.02$\end{tabular}} & \textbf{\begin{tabular}[c]{@{}l@{}}0.66\\ $\pm.02$\end{tabular}} \\ \hline
\end{tabular}

\label{ConflictDataset}
\end{table}

\begin{table}
\centering
\caption{BERT classification performance on the Emotion Dataset with and without DA, for LLM based methods Llama2-7B is used.}
\begin{tabular}{|l|l|l|l|l|l|}
\hline
Model & DA     & Acc                                                              & F1                                                               & R                                                                & P                                                                \\ \hline
BERT  & Orig   & \begin{tabular}[c]{@{}l@{}}0.25\\ $\pm.02$\end{tabular}          & \begin{tabular}[c]{@{}l@{}}0.24\\ $\pm.01$\end{tabular}          & \begin{tabular}[c]{@{}l@{}}0.25\\ $\pm.02$\end{tabular}          & \begin{tabular}[c]{@{}l@{}}0.26\\ $\pm.02$\end{tabular}          \\ \cline{2-6} 
      & EDA    & \begin{tabular}[c]{@{}l@{}}0.26\\ $\pm.02$\end{tabular}          & \begin{tabular}[c]{@{}l@{}}0.25\\ $\pm.03$\end{tabular}          & \begin{tabular}[c]{@{}l@{}}0.24\\ $\pm.03$\end{tabular}          & \begin{tabular}[c]{@{}l@{}}0.26\\ $\pm.03$\end{tabular}          \\ \cline{2-6} 
      & AugGPT & \begin{tabular}[c]{@{}l@{}}0.23\\ $\pm.02$\end{tabular}          & \begin{tabular}[c]{@{}l@{}}0.23\\ $\pm.02$\end{tabular}          & \begin{tabular}[c]{@{}l@{}}0.23\\ $\pm.02$\end{tabular}          & \begin{tabular}[c]{@{}l@{}}0.24\\ $\pm.02$\end{tabular}          \\ \cline{2-6} 
      & PMix   & \begin{tabular}[c]{@{}l@{}}0.24\\ $\pm.02$\end{tabular}          & \begin{tabular}[c]{@{}l@{}}0.24\\ $\pm.03$\end{tabular}          & \begin{tabular}[c]{@{}l@{}}0.24\\ $\pm.03$\end{tabular}          & \begin{tabular}[c]{@{}l@{}}0.26\\ $\pm.02$\end{tabular}          \\ \cline{2-6} 
      & PAug   & \textbf{\begin{tabular}[c]{@{}l@{}}0.27\\ $\pm.02$\end{tabular}} & \textbf{\begin{tabular}[c]{@{}l@{}}0.26\\ $\pm.02$\end{tabular}} & \textbf{\begin{tabular}[c]{@{}l@{}}0.27\\ $\pm.02$\end{tabular}} & \textbf{\begin{tabular}[c]{@{}l@{}}0.27\\ $\pm.02$\end{tabular}} \\ \hline
\end{tabular}

\label{EmotionDataset}
\end{table}
\subsection{Data Augmentation Effect on Classification Performance}

To answer \emph{RQ1}, we evaluated the classification results of CNN, DistilBERT, and BERT models trained on the original datasets, and synthetically generated data by our method PromptAug, AugGPT,~\cite{dai2023auggpt}, PromptMix~\cite{sahu2023promptmix}, EDA~\cite{wei2019eda}, and CBERT~\cite{wu2019conditional} DA methods.
Table \ref{ConflictDataset} shows the results, where Llama2-7B is used for LLM-based DA methods.
At the first glance, we notice PromptAug achieves the best performance, obtaining 2\% increases in accuracy and F1-score over the original dataset using the BERT classifier for Conflict dataset. This result is statistically significant using a T-test at p<0.05 with a t-value of 3.20 and a p-value of 0.03.Additionally, PromptAug is the only DA method that significantly outperforms the baseline dataset, achieving higher scores than other DA methods across all metrics, with the exception of AugGPT where precision is tied.
AugGPT outperforms the Orig baseline dataset but not significantly, with a t-value of 1.41 and a p-value of 0.23. 
For both CNN and DistilBERT classifiers, PromptAug outperforms both the original dataset and other augmentation models. The effects of all DA methods applied with the CNN model are very prominent, whilst the effects of DA are less evident but still present with DistilBERT. For the Emotion task in Table \ref{EmotionDataset}, PromptAug outperforms all models, with an increase of 2\% accuracy and 2\% F1-score over the original dataset. This result is significant at p <0.05, with a t-value of 3.09 and a p-value of 0.04. No other DA method achieved a significant improvement.

To assess generalizability of our DA method, we evaluated PromptAug and PromptMix with a different LLM, Mistral-8B. Here, we focus on BERT model due to it's superior performance. Table~\ref{LLM_comparison} presents the comparison between Llama- and Mistral-based DA methods.
We observe that PromptAug improves F1 score over the Original dataset baseline when using both Mistral-8B and Llama2-7B. AugGPT only achieved an increased F1 score using Llama2-7B, whilst PromptMix failed to improve F1 score using either LLM. We see that for both methods, Llama2-7B results in a stronger classification performance.

\begin{figure}[t]

\centering
\includegraphics[width=390pt]{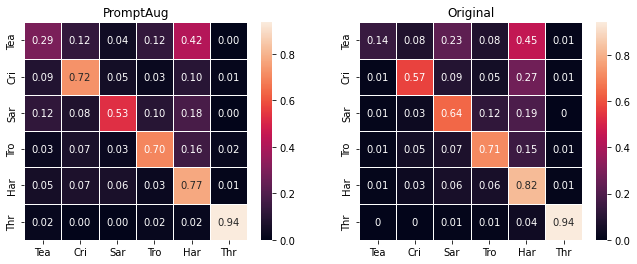}
\caption{Class Breakdown of BERT performance using the original dataset and PromptAug generated data.} \label{PromptHeat}
\end{figure}

\begin{table*}[ht!]
\centering
\caption{Comparison of LLM-based DA methods using Llama2-7B and Mistral-8B with BERT classifier.}
\begin{tabular}{|l|l|l|l|l|l|}
\hline
                                                 \multicolumn{2}{|c|}{} & Acc  & F1   & R  & P  \\ \hline
\multicolumn{2}{|c|}{Original Dataset} & 0.70 & 0.63 & 0.63 & 0.65     \\ \hline                     
\multicolumn{1}{|c|}{\multirow{3}{*}{Llama2}} 
    & PMix & 0.70 & 0.62 & 0.63 & 0.64
    \\ \cline{2-6} 
\multicolumn{1}{|c|}{}                           
    & AugGPT & 0.71 & 0.64 & 0.64 & \textbf{0.66} \\ \cline{2-6} 

\multicolumn{1}{|c|}{}  & PAug & \textbf{0.72} & \textbf{0.65} & \textbf{0.65} & \textbf{0.66}  \\ \hline
\multirow{3}{*}{Mistral}
    & PMix & 0.65 & 0.61 & 0.61 & 0.63\\ \cline{2-6} 
\multicolumn{1}{|c|}{}                           
    & AugGPT & \textbf{0.68} & 0.63 & \textbf{0.63} & 0.64 \\ \cline{2-6} 
\multicolumn{1}{|c|}{}  & PAug & 0.67 & \textbf{0.64} & \textbf{0.63} & \textbf{0.65}  \\ \hline
\end{tabular}

\label{LLM_comparison}
\end{table*}
 We further investigate the performance of the BERT classifier across different classes using the original dataset and PromptAug generated data.The results are shown in Figure~\ref{PromptHeat}.
 We observe large performance increases of 0.15 within `Teasing' and `Criticism' classes, marginal performance increase in `Trolling', and no performance increase for `Threat'. Despite an overall performance increase, there were performance decreases of 0.11 in `Sarcasm' and 0.05 in `Harassment'. In the original dataset `Teasing' and `Criticism' were most frequently misclassified as `Harassment'. This trend was reduced across almost all classes after DA. We propose that PromptAug increased these classes' profiles, reinforcing their identities as separate behaviours to `Harassment'. This highlights PromptAug's ability to be effective in scenarios with strong overlap between class boundaries and complex class behaviours. Class size could also be a contributing factor to performance increases, `Teasing' was by far the smallest and worst performing class, it therefore could have had the most to gain from an increase in datapoints and stature. PromptAug more than doubled the `Teasing' class performance, demonstrating it's effectiveness within a small, imbalanced multiclass dataset.

\textbf\emph{{Discussion.}}
Our results show PromptAug is an effective DA technique that can easily be used to improve classification performance. We highlight PromptAug's robustness and generalisability by showing improved performance over two datasets compared to two SOTA and two common DA methods, and evaluating the use of two different generative LLMs within the method. Additionally, the lack of pre-training and ease of access means that PromptAug maintains a simple approach. Enabling it's application to other tasks, only requiring an open source LLM and elements that researchers will already have when constructing datasets: task instruction and context, existing class examples, and class definitions. Although we do not conduct a quantitative efficiency analysis of the DA methods, PromptAug is more efficient than AugGPT and PromptMix. Both AugGPT and Promptmix require a model to be trained on the original dataset, PromptAug reduced computational cost by avoiding a pre-training step due to it's robust prompt desing and assertation mechanism. Additionally, AugGPT uses one original datapoint per LLM prompt whilst PromptAug uses three, therefore AugGPT requires more LLM calls to fully augment a dataset.

\subsection{Data Augmentation performance in Data Scarce Scenarios} 

To address \emph{RQ2}, we examine classification performance in simulated data scarcity scenarios, exploring how various DA methods can mitigate the impact of limited data availability.
DA techniques are frequently employed when there is a lack of available training data. Therefore, it is vital that the augmentation method retains its ability to create quality datapoints with limited data. DA methods which rely on a training step may struggle in extreme data scarcity scenario due to a lack of initial training data. Other issues are also worsened when data gets increasingly scarce; e.g., existing flaws in the dataset will become more pronounced if DA methods rely too heavily on the existing data. In order to mimic these data scenarios, we restrict the volume of training data available to the augmentation methods to 20\%, 40\%, 60\%, and 80\% of the original training dataset. This experiment demonstrates not only the effect of  training dataset size on classification models, but also the effectiveness of our augmentation method in data scarcity scenarios. It is vital to study how DA methods perform in these scenarios as it is a real-world representation of DA use cases.

\begin{figure}[t]

\centering
\includegraphics[width=390pt]{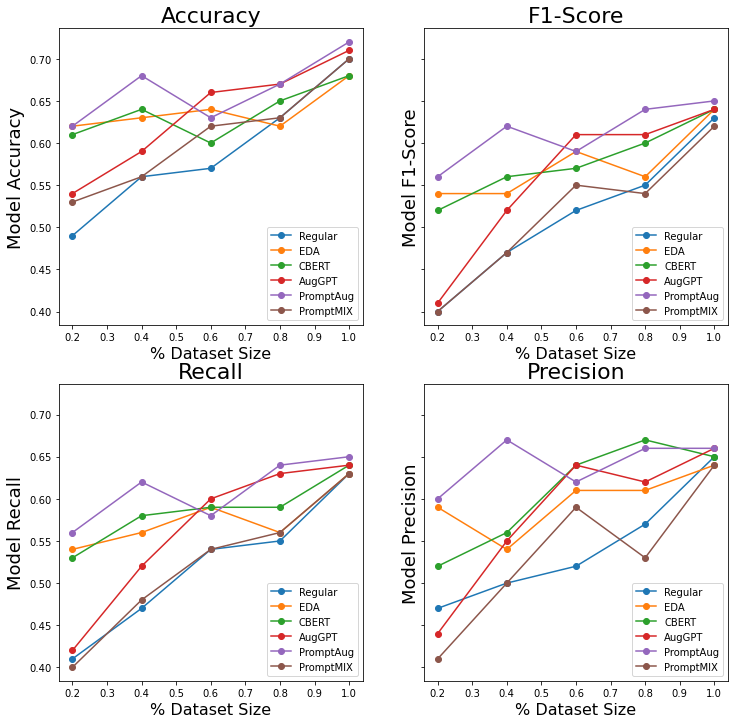}
\caption{Line graphs of performance vs dataset size.} \label{lines}
\end{figure}

Figure~\ref{lines} shows that for the original dataset, performance decreases as dataset size decreases. However, using the use of DA techniques, we can alleviate the issue. Of the methods, PromptAug improves the most over the original dataset, with accuracy increases of 13\%, 12\%, 6\%, 4\% and 2\% over dataset sizes of 20\%, 40\%, 60\%, 80\% and 100\%, respectively. This suggests that, for accuracy, DA is effective at all dataset sizes but has greater effect at lower dataset sizes. For F1-score, over the same size intervals, PromptAug improves over the baseline by 16\%, 15\%, 7\%, 9\%, and 2\%. PromptAug, therefore, has greater impact on F1-score compared to accuracy at most size intervals. 
PromptMix follows the same trend as the other DA methods with increased accuracy over the original dataset, but does not follow the trend of increased F1-scores. Although AugGPT outperforms PromptAug at 60\% it achieves smaller performance increases at the other four size intervals. Additionally, both AugGPT and PromptMix experience large dropoffs in performance increases at 20\% and 40\%. 

\textbf{\emph{Discussion.}}
Concluding experiment two, classification performance drops within the scenarios with the simulated smaller training datasets. This adverse effect can be mitigated using DA. PromptAug is the most effective DA technique, consistently increasing Accuracy and F1-score performance at all dataset sizes. We believe this is due to it's innovative approach, which provides rich contextual data with the instruction, context, definition, and example prompt components. This means that even with relatively few original datapoints that it can effectively generate quality data. By demonstrating PromptAug's ability to effectively operate in data scarce scenarios we show its suitability for DA, where tasks seeking to employ DA are frequently struggling with extreme data scarcity.

\subsection{ Quality Analysis of Augmented Datapoints} 

We answer RQ3 in two different ways: first, measuring diversity, and second by qualitatively analysing datapoints. 

\medskip
\textbf{\emph{Diversity Measure.}}
\citet{joko2024doing} highlight concerns about the diversity of augmented data points generated by LLM-based methods~\cite{joko2024doing, reif2023visualizing}. LLMs can be prone to repetition in generated data, and by using diversity metrics we can determine how similar the generated datapoints are to each other and the original datapoints. We employed two diversity metrics—Distinct-n (Dist-n)~\cite{li2015diversity} and Self-BLEU~\cite{zhu2018texygen}—to evaluate our results, following  the methodology  in \citet{joko2024doing}.
Dist-n evaluates the ratio of distinct n-grams to the total numbers of n-grams, respectively. A higher Dist-n score means that the dataset is more diverse. This can be interpreted in two ways when analysing augmented datapoints. Firstly, it is not good to have data that is so diverse that it deviates from the original characteristics of the dataset. Secondly, if diversity is very low, a classification model trained on that data may struggle to identify class datapoints outside the narrow representation in the augmented dataset. Self-BLEU examines the diversity present within a corpus by calculating the BLEU score between each datapoint in the corpus and the other datapoints in the corpus, taking an average of the scores. We use NLTK's BLEU methods to obtain each score, and set the weights for 1,2,3, and 4 n-grams to 0.25 each. In order to fully evaluate DA diversity, we calculate Self-BLEU diversity both within each DA methods' generated datapoints and between each DA methods generated datapoints and the datapoints in the original dataset. Following \citet{joko2024doing}, we applied normalization prior to computing both diversity metrics. We randomly sample from each set of augmented datapoints until a set number of words is reached. \citet{joko2024doing} note that without normalisation diversity metrics such as Dist-n are biased towards datasets with fewer words. 

The diversity analysis in Table~\ref{Diversity_Table} highlights that substitution based DA methods (EDA and CBERT) exhibit more diversity within the generated datapoints than LLM based methods over Dist-n and Self-BLEU within augmented data metrics. Potentially due to substitution based augmented datasets closely mirroring original datapoints, therefore retaining the original data's diversity. However, the metric "Self-BLEU comparing generated datapoints against original data" tells a different story, showing that data generated by the LLM methods AugGPT and PromptAug have higher diversity than those created via substitution methods. This can be attributed to LLM based methods creating entirely new datapoints rather than modifying existing datapoints.

\medskip
\textbf{\emph{Thematic Analysis.}}
To qualitatively analyse data quality we employ Thematic Analysis (TA), a widely used research method in the social science domain, formally established by \citet{braun2006using}. TA identifies and evaluates trends and patterns within a set of data, and for our paper we examined mis-annotated datapoints. This helped us to gain a greater understanding of the nuanced issues DA methods face, and in combination with the quantitative methods used above, it enhances the robustness of our paper's evaluation of DA methods. We sampled 150 datapoints from the augmented EDA and PromptAug data and conducted an independent annotation by two researchers, one of whom had no prior familiarity with the paper. One researcher coded the mis-annotated datapoints and identified themes, a second then reviewed the identified codes and themes. The researchers then discussed the codes, patterns, and themes before finalising findings, which are reported in the identified themes, definitions, descriptions, and examples included for robustness and reproducibility. We also conducted percentage annotator agreement and calculate Cohen's Kappa statistic according to \citet{mchugh2012interrater}.

\begin{table*}[ht!]
\centering
\caption{Diversity metrics for the  DA models. $\downarrow$ indicates a lower result is better. }
\begin{tabular}{ccc}
\hline
\multicolumn{1}{|c|}{}           & \multicolumn{1}{c|}{\textbf{Dist-1}}                                                                            & \multicolumn{1}{c|}{\textbf{Dist-2}}                                                                             \\ \hline
\multicolumn{1}{|c|}{EDA}        & \multicolumn{1}{c|}{\textbf{0.125}}                                                                             & \multicolumn{1}{c|}{\textbf{0.633}}                                                                              \\ \hline
\multicolumn{1}{|c|}{CBERT}      & \multicolumn{1}{c|}{0.104}                                                                                      & \multicolumn{1}{c|}{\textit{0.534}}                                                                              \\ \hline
\multicolumn{1}{|c|}{AugGPT}     & \multicolumn{1}{c|}{0.098}                                                                                      & \multicolumn{1}{c|}{0.393}                                                                                       \\ \hline
\multicolumn{1}{|c|}{Prompt Aug} & \multicolumn{1}{c|}{0.098}                                                                                      & \multicolumn{1}{c|}{0.439}                                                                                       \\ \hline
\multicolumn{1}{|c|}{Prompt Mix} & \multicolumn{1}{c|}{\textit{0.124}}                                                                             & \multicolumn{1}{c|}{0.419}                                                                                       \\ \hline
\multicolumn{1}{l}{}             & \multicolumn{1}{l}{}                                                                                            & \multicolumn{1}{l}{}                                                                                             \\ \hline
\multicolumn{1}{|l|}{}           & \multicolumn{1}{c|}{\textbf{\begin{tabular}[c]{@{}c@{}}Self-BLEU Within\\  Aug Data $\downarrow$\end{tabular}}} & \multicolumn{1}{c|}{\textbf{\begin{tabular}[c]{@{}c@{}}Self-BLEU Aug vs \\ Orig Data $\downarrow$\end{tabular}}} \\ \hline
\multicolumn{1}{|c|}{EDA}        & \multicolumn{1}{c|}{\textbf{0.132}}                                                                             & \multicolumn{1}{c|}{0.292}                                                                                       \\ \hline
\multicolumn{1}{|c|}{CBERT}      & \multicolumn{1}{c|}{\textit{0.150}}                                                                             & \multicolumn{1}{c|}{0.318}                                                                                       \\ \hline
\multicolumn{1}{|c|}{AugGPT}     & \multicolumn{1}{c|}{0.429}                                                                                      & \multicolumn{1}{c|}{\textbf{0.132}}                                                                              \\ \hline
\multicolumn{1}{|c|}{Prompt Aug} & \multicolumn{1}{c|}{0.479}                                                                                      & \multicolumn{1}{c|}{\textit{0.185}}                                                                              \\ \hline
\multicolumn{1}{|c|}{Prompt Mix} & \multicolumn{1}{c|}{0.452}                                                                                      & \multicolumn{1}{c|}{0.345}                                                                                       \\ \hline
\end{tabular}

\label{Diversity_Table}
\end{table*}

The thematic analysis performed on mis-annotated datapoints from EDA and PromptAug datasets produced four themes: (i) ``Linguistic Fluidity,'' (ii) ``Humour Ambiguity,'' (iii) ``Augmented Content Ambiguity,'' and (iv) ``Augmented Content Misinterpretation.'' 

The \emph{Linguistic Fluidity} theme encompasses fluid, blurred boundaries between classes. Although datapoints have dominant behaviours, they can contain aspects of multiple behaviours. \citet{jhaver2017view}  and \citet{kim2022does} investigate ambiguous class boundaries; i.e., how Criticism develops into Harassment, inter-relation between the two behaviors, and subjectivity of true class identity. Within hate research, \citet{fortuna2020toxic} also discuss how differing terminology in the domain leads to fluidity between behaviour classes in different works and misinterpretation of behavioural identities. 

The second theme, \emph{Humour Ambiguity}, relates to the difficulty of identifying nuanced forms of humour. Humour has been recognised as a challenging NLP area. It is largely subjective and often relies on subtle cues. For example, the first humour ambiguity datapoint in Table \ref{Coding_table} belongs to 'Trolling' but was mis-annotated as `Teasing'. There are two difficulties in identifying this datapoint. Firstly, the border between teasing and trolling behaviours can be subjective, in that what one individual finds humourous may come across as offensive to another. Secondly, altering datapoints using DA may worsen the already ambiguous and blurred class boundaries making annotation more difficult.

The third theme, \emph{Augmented Content Ambiguity}, relates to the ability of DA methods to generate coherent and interpretable datapoints whilst retaining class labels. For complex interaction behaviours, class labels can depend on subtle text features, and DA methods can obscure and remove vital clues for human coders. In the two given examples, one can observe that text transmutation has compromised the sentence composition, resulting in difficult interpretation for human coders. \citet{chen2023empirical} note a similar problem of text transmutation changing the meaning of sentences in their NLP DA survey.

The final theme, \emph{Augmented Content Misinterpretation}, occurs within the PromptAug data. Although designed to produce quality dataset examples, PromptAug occasionally produces erroneous responses, e.g., other negative classes, advice on dealing with the behaviour, and random data. These responses are difficult to filter and do not reflect the desired class,  hindering model performance. These erroneous responses are often a result of LLM safety nets used to ensure safe AI practices. \citet{lermen2023lora} identify this issue when generating negative behaviour datapoints related to this paper's domain, finding that Llama can refuse to produce conflict comments examples such as harassment and hate  around 75\% and 70\% of the time.

Both EDA and PromptAug methods experienced Linguistic Fluidity and Humour Ambiguity. Augmented Content Ambiguity was identified in EDA's data, and Augmented Content Misinterpretation was identified in PromptAug's data (Tables.~\ref{Coding_table},\ref{Coding_table2}). For PromptAug, data annotators had an agreement rate of 67\% and Cohen's K of 0.36, described as "fair agreement" by \citet{landis1977measurement}. For EDA, data annotators had an agreement rate of 46\% and Cohen's K of 0.14, described as ``slight agreement." Identifying thematic analysis themes provides an evaluation of DA beyond quantitative metrics. These themes can be used to target weaknesses found in all NLP DA methods such as linguistic fluidity and humour ambiguity, or used to target specific weaknesses such as augmented content ambiguity for EDA or augmented content misinterpretation for PromptAug.

\subsection{Effect of PromptAug's 
Components on classification}
To answer \emph{RQ4}, we investigate how the different components of PromptAug combine to produce an effective DA technique. We conduct an ablation study with both the emotion and conflict datasets, evaluating PromptAug without one of it's components at each setting. For the examples and defintion PromptComponents we remove them in their entirety. As we could not entirely remove the instruction component, otherwise no datapoints would be generated, we instead combine the removal of some of the instruction component with the removal of the context component.

We can see from Table \ref{component_ablation}, PromptAug performs the best in both the conflict and emotion datasets when all of it's components are present. Within the conflict dataset we see that ``no examples'' and ``no definition'' variants still outperforming the baseline, although not achieving the same performance as the full PromptAug method. The ``no context'' variant is the only one which lowers performance compared to the baseline conflict dataset performance. Within the emotion dataset there is a different story. Although the full PromptAug method improves performance over the baseline, all of the variants actually decrease performance compared to the baseline dataset.

\begin{table}[ht!]
\centering
\caption{Results of the Prompt Component ablation study on classification effectiveness.}

\begin{tabular}{|l|llll|l|llll|}
\hline
                                                                  & \multicolumn{4}{l|}{Conflict Dataset}                                                    &  & \multicolumn{4}{l|}{Emotion Dataset}                                                     \\ \cline{1-5} \cline{7-10} 
DA                                                                & \multicolumn{1}{l|}{Acc}  & \multicolumn{1}{l|}{F1}   & \multicolumn{1}{l|}{Rec}  & Pre  &  & \multicolumn{1}{l|}{Acc}  & \multicolumn{1}{l|}{F1}   & \multicolumn{1}{l|}{Rec}  & Pre  \\ \cline{1-5} \cline{7-10} 
Original                                                          & \multicolumn{1}{l|}{0.70} & \multicolumn{1}{l|}{0.63} & \multicolumn{1}{l|}{0.63} & 0.65 &  & \multicolumn{1}{l|}{0.25} & \multicolumn{1}{l|}{0.24} & \multicolumn{1}{l|}{0.25} & 0.26 \\ \cline{1-5} \cline{7-10} 
PromptAug                                                         & \multicolumn{1}{l|}{0.72} & \multicolumn{1}{l|}{0.65} & \multicolumn{1}{l|}{0.65} & 0.66 &  & \multicolumn{1}{l|}{0.27} & \multicolumn{1}{l|}{0.26} & \multicolumn{1}{l|}{0.27} & 0.27 \\ \cline{1-5} \cline{7-10} 
\begin{tabular}[c]{@{}l@{}}PromptAug\\ No Examples\end{tabular}   & \multicolumn{1}{l|}{0.72} & \multicolumn{1}{l|}{0.64} & \multicolumn{1}{l|}{0.64} & 0.66 &  & \multicolumn{1}{l|}{0.23} & \multicolumn{1}{l|}{0.23} & \multicolumn{1}{l|}{0.24} & 0.24 \\ \cline{1-5} \cline{7-10} 
\begin{tabular}[c]{@{}l@{}}PromptAug\\ No Definition\end{tabular} & \multicolumn{1}{l|}{0.71} & \multicolumn{1}{l|}{0.64} & \multicolumn{1}{l|}{0.62} & 0.69 &  & \multicolumn{1}{l|}{0.23} & \multicolumn{1}{l|}{0.23} & \multicolumn{1}{l|}{0.24} & 0.26 \\ \cline{1-5} \cline{7-10} 
\begin{tabular}[c]{@{}l@{}}PromptAug\\ No Context\end{tabular}    & \multicolumn{1}{l|}{0.70} & \multicolumn{1}{l|}{0.62} & \multicolumn{1}{l|}{0.61} & 0.67 &  & \multicolumn{1}{l|}{0.23} & \multicolumn{1}{l|}{0.23} & \multicolumn{1}{l|}{0.24} & 0.26 \\ \hline
\end{tabular}
\label{component_ablation}
\end{table}
\textbf{Discussion}
The full PromptAug variant with all prompt components performing the best was to be expected, we carefully design the method as described in the methodology so that each component represents an important contribution. Two surprises occurred in the other results. Firstly, that when removing prompt components PromptAug performed better in the conflict dataset than the emotion dataset. Secondly, that all variants within the emotion dataset decreased performance. We had thought that the emotion dataset would perform better in the variants as the behaviours are more standardised compared to the conflict dataset. Perhaps without all prompt components present the emotion dataset suffered more from the issues identified with DA techniques in the thematic analysis above. Concluding this experiment we can see that removing any one element of PromptAug decreases overall performance, and that only by utilising them all together do we see the ability of the technique to improve classification performance.

\subsection{Overall Discussion}
In this section, we have evaluated PromptAug and other DA methods in an extensive and robust manner. Ultimately the goal of a DA method should be to increase classification performance. We have shown PromptAug's ability to not just increase the standard model performance metrics but its ability across an extended suite of quantitative and qualitative methods. Within RQ1, we show how PromptAug can improve classification performance across a range of classification models and datasets using different generative LLMs within the method. RQ2, investigates DA performance in extreme data scarce scenarios, a vital concern of LLM data generation that frequently goes unaddressed in research papers. We are able to show that PromptAug is not only effective in data scarce scenarios but also outperforms the other methods significantly. RQ3 examines qualitative and quantitative quality of generated datapoints. Conducting either a thematic analysis or diversity analysis is not commonly done in DA papers, let alone including both for a more rigorous analysis. Here we show that we have conducted an honest and thorough evaluation of PromptAug: while it performs well in some of this experiments metrics it is not the outright top performer across all metrics. It is however competitive even in those that it isn't the best. Finally, in order to reinforce the novelty and to support the methodology we use an ablation study over PromptAug's components.

\subsection{Limitations}
We evaluate our model's generalisability across various classification models, datasets, and data scarce scenarios. However, both datasets included had small class sizes for 6 and 11 multiclass scenarios. Therefore, we cannot assume the generalisation of our method to other datasets with two or three classes or those with large class sizes, as they might not suffer the issues DA aims to solve. Furthermore, we evaluate on two human behaviour related datasets, the other SOTA methods may not struggle as much when datapoints do not consist of nuanced human behaviour. Additionally, we only use Llama-7B and Mistral-7B as generative LLMs for our method, so we cannot assume any generalisability for more powerful LLMs such as GPT-4 or GPT-3.5 turbo. We also do not investigate any social bias present within the datapoints generated by the LLM.

\subsection{Ethical Concerns}
In this paper we discuss harmful content, e.g., harassment and threats, and how to generate it using LLMs. This presents an opportunity for individuals with malicious intent to use this research to cause harm. We argue that the purpose behind this work is to improve the classification performance of harmful content along a negative behaviour spectrum. This increased ability to successfully identify harmful content on social media is ultimately a net positive for society. In addition, we don't specify any additional techniques to completely bypass LLMs safety nets, instead we only note that our prompt structure does do so to some degree.

\section{Conclusion}
We present a novel two step DA approach based on LLM prompting, targeting class definition and identity within two small, complex human behaviour multi-class datasets. Our augmentation method harnesses the power of LLMs while being easily implemented, requiring no finetuning, and achieving superior classification performance over two baseline datasets and other SOTA DA methods. We further demonstrate the effectiveness of the augmentation method in extreme data scarce scenarios. We quantitatively analyse the quality of the generated data by evaluating diversity within augmented datapoints. In addition, we conduct a qualitative thematic analysis of the augmented datapoints finding within augmented datapoints there are four main themes of mis-annotation; linguistic fluidity, humour ambiguity, augmented content ambiguity, and augmented content misinterpretation.

With recent emphasis on responsible AI and a growing focus on social bias within LLMs, future work could examine how bias presents itself within DA. A study adopting two methods suggested by \citet{ferrara2023should}, `Applying fairness metrics' and `Human-in-the-loop approaches', would provide insights on social bias of generated data. Secondly, quantifying expenses of DA methods would be of interest, highlighting trade-offs between expense and performance. Future work could also employ PromptAug within other text datasets or use more powerful LLMs within the method, evaluating generalisability.
\bibliographystyle{ACM-Reference-Format}

\appendix

\renewcommand{\thesection}{\Alph{section}}

\section{Appendix}
\begin{sidewaystable*}
\centering
\caption{Tables showing package versions and URLs.}

\begin{tabular}{lll}
\hline
\multicolumn{1}{|l|}{\textbf{Package}} & \multicolumn{1}{l|}{\textbf{Version}} & \multicolumn{1}{l|}{\textbf{URL}}                                 \\ \hline
\multicolumn{1}{|l|}{Huggingface Hub}  & \multicolumn{1}{l|}{0.20.3}           & \multicolumn{1}{l|}{https://huggingface.co/}                \\ \hline
\multicolumn{1}{|l|}{Accelerate}       & \multicolumn{1}{l|}{0.26.1}           & \multicolumn{1}{l|}{https://huggingface.co/docs/accelerate} \\ \hline
\multicolumn{1}{|l|}{Transformers}     & \multicolumn{1}{l|}{4.35.2}           & \multicolumn{1}{l|}{https://huggingface.co/docs/transformers/}    \\ \hline
\multicolumn{1}{|l|}{Torch}            & \multicolumn{1}{l|}{2.2.0}            & \multicolumn{1}{l|}{https://pypi.org/project/torch/}              \\ \hline
\multicolumn{1}{|l|}{Pandas}           & \multicolumn{1}{l|}{1.5.3}            & \multicolumn{1}{l|}{https://pandas.pydata.org/}                   \\ \hline
\multicolumn{1}{|l|}{Numpy}            & \multicolumn{1}{l|}{1.25.2}           & \multicolumn{1}{l|}{https://numpy.org/}                           \\ \hline
\multicolumn{1}{|l|}{Sklearn}          & \multicolumn{1}{l|}{1.4.1}            & \multicolumn{1}{l|}{https://scikit-learn.org/stable/}             \\ \hline
\multicolumn{1}{|l|}{Meta Llama}       & \multicolumn{1}{l|}{Llama-2-7b}       & \multicolumn{1}{l|}{https://huggingface.co/meta-Llama}            \\ \hline
                                       &                                       &                                                                  
\end{tabular}
\label{packages}
\end{sidewaystable*}

\begin{sidewaystable*}
\caption{Table showing an example prompt and it's outputs.}
\centering
\begin{tabular}{|c|l|}
\hline
Prompt    & \begin{tabular}[c]{@{}l@{}}In a numbered list, write 5 new social media comments containing\\ Sarcasm directed at other users. Here are some examples;\\ \\ "oh dear boo Hoo",\\ "Wow, you've sure convinced me a bill of rights drafted by British\\ representatives elected by the British public which can be changed\\ at anytime should the British people deem it necessary is a bad\\ idea",\\ "Sounds like a stable police officer."\\ \\ Sarcasm is defined as; humorous communication (bitter, biting,\\ cynical, hurtful tone, incl. swearwords)\end{tabular} \\ \hline
Responses & \begin{tabular}[c]{@{}l@{}}1. "Oh wow, your opinion on climate change is as valid as a Greta\\ Thunberg tweet "\\ 2. "Thanks for sharing your expertise on quantum physics, I'm\\ sure Einstein is rolling in  his grave"\\ 3. "I'm so glad you're here to teach me about the importance of\\ recycling, I was totally  unaware of the impact of my trash on the\\ environment "\\ 4. "Your argument is as solid as a Trump tweet "\\ 5. "I'm glad you're passionate about your beliefs, but maybe try\\ using facts instead of insults "\end{tabular}                    \\ \hline
\end{tabular}
\label{Exampleprompt}
\end{sidewaystable*}

\begin{sidewaystable*}
\centering
\caption{Linguistic Fluidity and Humour Ambiguity themes identified in Thematic Analysis of DA datapoints.}
\begin{tabular}{|l|l|l|l|}
\hline
\multicolumn{1}{|c|}{\textbf{Theme}}                                                             & \multicolumn{1}{c|}{\textbf{Definition}}                                                                                                                                                                                                     & \multicolumn{1}{c|}{\textbf{Description}}                                                                                                                                                                                                                                                                                                                                                                                                                                                    & \multicolumn{1}{c|}{\textbf{Examples}}                                                                                                                                                                                                                                                                                                                                                                    \\ \hline
\textbf{\begin{tabular}[c]{@{}l@{}}Linguis-\\ tic \\ Fluidity\end{tabular}}                      & \begin{tabular}[c]{@{}l@{}}A miscoding of an  
 \\ augmented datapoint\\ that occurs due to the\\ lack of defintional\\ boundaries that are\\ inherent to the\\ interpretation of\\ language.\end{tabular}                                         & \begin{tabular}[c]{@{}l@{}}A commonly known phenomenon in \\ linguistics is that of multiple meanings\\ to the same sentence, where\\ interpretation depends on a multitude \\ of unpredictable factors(e.g. one's mood\\ , need for politeness etc;) Classes are \\ not always clear cut, often having \\ fluid boundaries. Datapoints can \\ contain behaviour which could belong\\  to more than one class, making it\\difficult for annotators to get it totally\\ accurate. \end{tabular} & \begin{tabular}[c]{@{}l@{}}Coded - "Harassment, \\ Actual Class - "Sarcasm"\\ "I'm not sure what's more\\ impressive: your ability to\\ take a selfie or your lack\\ of self-awareness..."\\ \\ Coded - "Harassment, \\ Actual Class - "Trolling"\\ "I can't stand this\\ YouTuber's voice. It's\\ like fingernails on a\\ chalkboard
every time\\ they speak."\\\end{tabular}                                      \\ \hline
\textbf{\begin{tabular}[c]{@{}l@{}}Humour\\ Ambig-\\ uity\end{tabular}}                          & \begin{tabular}[c]{@{}l@{}}A miscoding of an \\ augmented datapoint\\ that occurs when a\\ message fails to convey\\ that it was meant in\\ humour and/or  was\\ good vs. bad-natured.\end{tabular}                                             & \begin{tabular}[c]{@{}l@{}}Linguists have long recognised the lack\\ of clarity inherent to humour as a \\ quality on which humour often relied.\\  Humour has been recognised as a \\ particularly challenging area of NLP. \\ Humour can often be taken two ways \\ and is subjective meaning the \\ dominant type of humour behaviour \\
is often ambiguous within datapoints.\\\end{tabular}                                                                                                 & \begin{tabular}[c]{@{}l@{}}Coded - "Teasing", \\ Actual Class - "Trolling" \\ "Your favorite meme is\\ so last year, get with the\\ times"\\ \\ Coded - "Teasing", \\ Actual Class - "Trolling"\\ "he makes that phone\\ look like
a tablet"\\\end{tabular}                                                                                                                                                    \\ \hline

\end{tabular}

\label{Coding_table}
\end{sidewaystable*}

\begin{sidewaystable*}
\centering
\caption{Augmented Content Ambiguity and Misinterpretation themes identified in Thematic Analysis of DA datapoints.}
\begin{tabular}{|l|l|l|l|}
\hline
\multicolumn{1}{|c|}{\textbf{Theme}}                                                             & \multicolumn{1}{c|}{\textbf{Definition}}                                                                                                                                                                                                     & \multicolumn{1}{c|}{\textbf{Description}}                                                                                                                                                                                                                                                                                                                                                                                                                                                    & \multicolumn{1}{c|}{\textbf{Examples}}                                                                                                                                                                                                                                                                                                                                                                    \\ \hline
\textbf{\begin{tabular}[c]{@{}l@{}}Augmen-\\ ted\\ Content\\ Ambig-\\ uity\end{tabular}}         & \begin{tabular}[c]{@{}l@{}}A miscoding of an \\ augmented datapoint\\ that occurs due to a\\ lack of clarity within\\ the datapoint\\ produced by the \\ augmentation\\ technique,  where the\\ content makes no \\ coherent sense.\end{tabular} & \begin{tabular}[c]{@{}l@{}}Within NLP DA label preservation is a \\ known challenge, where class boundaries \\ can depend on specific and nuanced words, \\ phrases, and subtleties. Text transmutation \\ such as synonym swapping/insertion, word \\ deletion, and reordering can change the \\ context and legibility of datapoints, severely \\ impacting label and datapoint behaviour \\ preservation.\end{tabular}                                                                    & \begin{tabular}[c]{@{}l@{}}Coded - "Harassment", \\ Actual Class - "Criticism"\\ "ua warrior same if probably \\steph had the of  type won\\ commercial for would've"\\ \\ Coded - "Trolling", \\ Actual Class - "Harassment"\\ "do you rattling have sex \\microsoft i dont  believe you\\ have sex what are you speak \\about"\end{tabular}                                                                   \\ \hline
\textbf{\begin{tabular}[c]{@{}l@{}}Augmen-\\ ted\\ Content\\ Misinter-\\ pretation\end{tabular}} & \begin{tabular}[c]{@{}l@{}}A miscoding of an \\ augmented  datapoint\\ that occurs due to\\ the augmentation\\ technique \\misinterpreting the \\ augmentation task.\end{tabular}                                                               & \begin{tabular}[c]{@{}l@{}}Although LLMs can be given specific \\ prompt instructions they do not always \\ generate datapoints within the specified \\ boundaries. Occasionally, instead of \\ generating examples of the requested \\ behaviour the LLM would instead \\ produce examples of responses to that \\ behaviour. These erroneous examples tend \\ not to adhere to the characteristics of the \\ class behaviour and can vary drastically \\ in their identity.\end{tabular}   & \begin{tabular}[c]{@{}l@{}}Coded - "Criticism", \\ Actual Class - "Trolling"\\ "Lol @ you thinking you're \\ relevant, get a life troll"\\ \\ Coded - "Criticism",\\ Actual Class -"Trolling"\\ "I'm so tired of @username \\ constantly posting memes\\
 that are offensive and\\ disrespectful. Can't they \\see how their humor is\\ affecting others?\\ 
\#harassment \#block"\\\end{tabular} \\ \hline
\end{tabular}

\label{Coding_table2}
\end{sidewaystable*}

\end{document}